\documentclass[11pt,a4paper]{article}
\usepackage{authblk}
\usepackage[hyperref]{acl2019}
\usepackage{times}
\usepackage{latexsym}
\usepackage{booktabs}
\usepackage{graphicx}
\usepackage{xcolor}
\usepackage{pgfplots}

\usepackage{url}

\usepackage{tikz}
\usepackage{booktabs}
\usepackage[normalem]{ulem}
\usepackage{amsmath}
\useunder{\uline}{\ul}{}

\newcommand{\DEVELOPMENT}{1} 

\usepackage{ifthen}
\ifthenelse{\DEVELOPMENT = 1}{
  \usepackage{todonotes}

}{

}

\newcommand{\fem}[1]{{\bf \color{purple} #1}}
\newcommand{\masc}[1]{{\bf \color{blue} #1}}
\newcommand{\neut}[1]{{\bf \color{orange} #1}}
\newcommand{\link}[0]{\url{shorturl.at/dimuD}}
\newcommand{\ds}[0]{WinoMT}
\newcommand{\wg}[0]{Winogender}
\newcommand{\wb}[0]{WinoBias}

\newcommand{\amazon}[0]{Amazon Translate}

\newcommand{\goog}[0]{Google Translate}
\newcommand{\bing}[0]{Microsoft Translator}
\newcommand{\systran}[0]{SYSTRAN}
\newcommand{\aws}[0]{Amazon Translate}

\aclfinalcopy 
\def\aclpaperid{2100} 

\title{Evaluating Gender Bias in Machine Translation}

\author[1,2]{\textbf{Gabriel Stanovsky}}
\author[1,2]{\textbf{Noah A. Smith}}
\author[1]{\textbf{Luke Zettlemoyer}}
 \affil[1]{Paul G.~Allen School of Computer Science \& Engineering, University of Washington, Seattle, USA}
 \affil[2]{Allen Institute for Artificial Intelligence, Seattle, USA}

\affil[  ]{\tt \{gabis,nasmith,lsz\}@cs.washington.edu}

\date{}

\begin{document}
\maketitle
\begin{abstract}
  We present the first challenge set and evaluation protocol for the analysis of gender bias in machine translation (MT). Our approach uses two recent coreference resolution datasets composed of English sentences which cast participants into non-stereotypical gender roles (e.g., ``The doctor asked the nurse to help \emph{her} in the operation''). 
  We devise an automatic gender bias evaluation method for eight target languages with grammatical gender, based on 
  morphological analysis (e.g., the use of female inflection for the word ``doctor'').
  Our analyses show that four popular industrial MT systems and two recent state-of-the-art academic MT models are significantly prone to gender-biased translation errors for all tested target languages.
  Our data and code are publicly available at \link.
\end{abstract}

\section{Introduction}
Learned models exhibit social bias when their training data encode stereotypes not relevant for the task, but the correlations are picked up anyway.
Notable examples include gender biases
in visual SRL (cooking is stereotypically done by women, construction workers are stereotypically men; \citealp{Zhao2017MenAL}),
lexical semantics (``man is to computer programmer as woman is to homemaker'';
\citealp{Bolukbasi2016ManIT}), and natural language inference (associating
women with gossiping and men with guitars; \citealp{Rudinger2017SocialBI}).

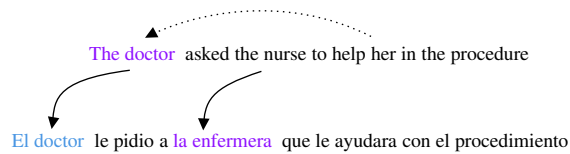
\begin{figure}[t!]
  \resizebox{\columnwidth}{!}{

\tikzset{every picture/.style={line width=0.75pt}} 

\begin{tikzpicture}[x=0.75pt,y=0.75pt,yscale=-1,xscale=1]

\draw    (212.25,91.18) .. controls (152.42,98.9) and (157.08,115.76) .. (154.78,133.11) ;
\draw [shift={(154.5,135)}, rotate = 279.46] [fill={rgb, 255:red, 0; green, 0; blue, 0 }  ][line width=0.75]  [draw opacity=0] (8.93,-4.29) -- (0,0) -- (8.93,4.29) -- cycle    ;

\draw    (309.5,92) .. controls (270.7,104.61) and (267.65,114.4) .. (266.59,134.14) ;
\draw [shift={(266.5,136)}, rotate = 272.73] [fill={rgb, 255:red, 0; green, 0; blue, 0 }  ][line width=0.75]  [draw opacity=0] (8.93,-4.29) -- (0,0) -- (8.93,4.29) -- cycle    ;

\draw  [dash pattern={on 0.84pt off 2.51pt}]  (390.5,66) .. controls (326.47,33.49) and (259.54,48.53) .. (225.05,66.19) ;
\draw [shift={(223.5,67)}, rotate = 332.1] [fill={rgb, 255:red, 0; green, 0; blue, 0 }  ][line width=0.75]  [draw opacity=0] (8.93,-4.29) -- (0,0) -- (8.93,4.29) -- cycle    ;

\draw (345,80) node [scale=1,color={rgb, 255:red, 0; green, 0; blue, 0 }  ,opacity=1 ]  {$\text{\textcolor[rgb]{0.56,0.07,1}{The\ doctor} \ asked\ the\ nurse\ to\ help\ her\ in\ the\ procedure}$};
\draw (331,145) node [scale=1,color={rgb, 255:red, 0; green, 0; blue, 0 }  ,opacity=1 ]  {$\text{\textcolor[rgb]{0.29,0.59,0.89}{El\ doctor} \ le\ pidio\ a\ \textcolor[rgb]{0.56,0.07,1}{la\ enfermera} \ que\ le\ ayudara\ con\ el\ procedimiento}$};

\end{tikzpicture}
}
    \caption{
      An example of gender bias in machine translation from English (top) to Spanish (bottom).
      In the English source sentence, the nurse's gender is unknown,
      while the coreference link with ``her'' identifies the ``doctor'' as a female.
      On the other hand, the Spanish target sentence uses morphological features for gender: ``\emph{el} doctor'' (male), versus ``\emph{la} enfermer\emph{-a}'' (female).
      Aligning between source and target sentences reveals that a stereotypical assignment of gender roles changed the meaning of the translated sentence by changing
      the doctor's gender.
      }

  \label{fig:flow}
\end{figure}

In this work, we conduct the first large-scale multilingual evaluation of gender-bias in machine translation (MT), following
recent small-scale qualitative studies
which observed that online MT services,
such as \goog\ or \bing, also exhibit biases,
e.g., translating nurses as females and
programmers as males, regardless of context \cite{AlvarezMelis2017ACF,DBLP:journals/corr/abs-1901-03116}.
\goog\ recently tried to mitigate these biases by allowing users to sometimes choose between gendered translations \cite{googleGenderMT}.
%




As shown in Figure \ref{fig:flow}, we use data introduced by two recent coreference gender-bias studies:
the \wg\ \cite{Rudinger2018GenderBI}, and the \wb\ \cite{Zhao2018GenderBI} datasets.
Following the Winograd schema \cite{Levesque2011TheWS}, each instance in
these datasets 
is an English sentence which describes a scenario with human entities, who are
identified by their role (e.g., ``the doctor'' and ``the nurse''
in Figure \ref{fig:flow}), and a
pronoun (``her'' in the example), which needs to be correctly resolved to one of the
entities (``the doctor'' in this case).
\newcite{Rudinger2018GenderBI} and \newcite{Zhao2018GenderBI} found that while human agreement on the task was high (roughly 95\%),
coreference resolution models
often ignore context and make socially biased predictions, e.g., associating the feminine pronoun ``her'' with the stereotypically female ``nurse.''


We observe that for many target languages, a faithful translation requires
a similar form of (at least implicit) gender identification.
In addition, in the many languages which associate between biological
and grammatical gender (e.g., most Romance, Germanic, Slavic, and Semitic languages; \citealp{craig1986noun,mucchi2005visible,corbett2007gender}),
the gender of an animate object can be identified via morphological markers.
For instance, when translating our running
example in Figure \ref{fig:flow} to Spanish,
a valid translation may be: ``\emph{La doctora} le pidio a  la enfermera  que le ayudara con el procedimiento,'' which indicates
that the doctor is a woman, by using a feminine suffix inflection (``doctor\emph{a}'') and
the feminine definite gendered article (``\emph{la}'').
However, a biased translation system may ignore the given context and stereotypically translate the doctor as male, as shown at the bottom of the figure.

Following these observations, we design a challenge set
approach for evaluating gender bias
in MT using a concatenation of \wg\ and \wb. 
We devise an automatic translation evaluation method for eight diverse target languages,
without requiring additional gold translations, relying instead on automatic measures for alignment and morphological analysis
(Section \ref{sec:bias}).
We find that four widely used commercial MT systems and two recent state-of-the-art academic models
are significantly gender-biased on all tested languages (Section \ref{sec:eval}).
Our method and benchmarks are publicly available, and are easily
extensible with more languages and MT models.

\section{Challenge Set for Gender Bias in MT}
\label{sec:bias}

We compose a challenge set for gender bias in MT (which we dub ``\ds'') by concatenating the \wg\ and \wb\ coreference test sets.
Overall, \ds\ contains 3,888 instances, and is equally balanced between male and female genders, as well as between stereotypical and non-stereotypical gender-role assignments
(e.g., a female doctor versus a female nurse).
Additional dataset statistics are presented in Table \ref{tab:dataset}.


We use \ds\ to estimate the gender-bias of an MT model, $M$, in target-language $L$ by performing following steps (exemplified in Figure \ref{fig:flow}):

\noindent (1) \textbf{Translate} all of the sentences in \ds\ into $L$ using $M$, thus forming a bilingual corpus of English and the target language $L$.

\noindent (2) \textbf{Align} between the source and target translations, using \emph{fast\_align}
  \cite{Dyer2013ASF}, trained on the automatic translations from from step (1).
  We then map the English entity annotated in the coreference datasets
  to its translation (e.g., align between ``the doctor'' and ``el doctor''
  in Figure \ref{fig:flow}).

\noindent (3) Finally,  we
  \textbf{extract the target-side entity's
  gender} using simple heuristics over
  language-specific morphological analysis, which we
  perform using off-the-shelf tools for each target language, as discussed in the following section.

This process extracts the translated genders, according to $M$, for all of the
entities in \ds, which we can then evaluate against the gold
annotations provided by the original English dataset.

This process can introduce noise into our evaluation in steps (2) and (3), via wrong alignments or erroneous morphological analysis.
In Section \ref{sec:eval}, we will present a human evaluation showing these errors are infrequent. 

\begin{table}[t!]
  \centering

    \begin{tabular}{@{}lrrr@{}}
      \toprule
      & \multicolumn{1}{c}{\small{\wg}} & \multicolumn{1}{c}{\small{\wb}} & \multicolumn{1}{c}{\small{\ds}} \\ \midrule
      Male           & 240                            & 1582                         & 1826                    \\
      Female         & 240                            & 1586                         & 1822                    \\
      Neutral        & 240                            & 0                            & 240                     \\
      \textbf{Total} & \textbf{720}                   & \textbf{3168}                & \textbf{3888}           \\ \bottomrule
    \end{tabular}%
  \caption{The coreference test sets and resulting \ds\
    corpus statistics (in number of instances).}
  \label{tab:dataset}
  \end{table}

\section{Evaluation}
\label{sec:eval}

\begin{table*}[t!]
  \centering
  \begin{tabular}{@{}lrrrrrrrrrrrr@{}}
    \toprule
    & \multicolumn{3}{c}{\small{\textbf{\goog}}}                                                                & \multicolumn{3}{c}{\small{\textbf{\bing}}}                                                                & \multicolumn{3}{c}{\small{\textbf{\aws}$^{*}$}}                                                                 & \multicolumn{3}{c}{\small{\textbf{\systran}}}                                                             \\
    & \multicolumn{1}{l}{\small{Acc}} & \multicolumn{1}{l}{$\Delta_G$} & \multicolumn{1}{l}{$\Delta_S$} & \multicolumn{1}{l}{\small{Acc}} & \multicolumn{1}{l}{$\Delta_G$} & \multicolumn{1}{r}{$\Delta_S$} & \multicolumn{1}{l}{\small{Acc}} & \multicolumn{1}{l}{$\Delta_G$} & \multicolumn{1}{l}{$\Delta_S$} & \multicolumn{1}{c}{\small{Acc}} & \multicolumn{1}{l}{$\Delta_G$} & \multicolumn{1}{c}{$\Delta_S$} \\ \midrule
    \multicolumn{1}{l|}{\textit{ES}} & 53.1                            & 23.4                           & \multicolumn{1}{r|}{21.3}      & 47.3                      & 36.8      & \multicolumn{1}{r|}{23.2}      & \textbf{59.4}             & 15.4                           & \multicolumn{1}{r|}{22.3}      & 45.6                            & 46.3                           & 15.0                           \\
    \multicolumn{1}{l|}{\textit{FR}} & {\ul \textbf{63.6}}             & 6.4                            & \multicolumn{1}{r|}{26.7}      & 44.7                            & 36.4      & \multicolumn{1}{r|}{29.7}      & 55.2                            & 17.7                           & \multicolumn{1}{r|}{24.9}      & 45.0                            & 44.0                           & 9.4                            \\
    \multicolumn{1}{l|}{\textit{IT}} & 39.6                            & 32.9                           & \multicolumn{1}{r|}{21.5}      & 39.8                            & 39.8      & \multicolumn{1}{r|}{17.0}      & \textbf{42.4}                   & 27.8                           & \multicolumn{1}{r|}{18.5}      & 38.9                            & 47.5                           & 9.4                            \\ \midrule
    \multicolumn{1}{r|}{\textit{RU}} & 37.7                            & 36.8                           & \multicolumn{1}{r|}{11.4}      & 36.8                            & 42.1      & \multicolumn{1}{r|}{8.5}       & \textbf{39.7}                   & 34.7                           & \multicolumn{1}{r|}{9.2}       & 37.3                            & 44.1                           & 9.3                            \\
    \multicolumn{1}{r|}{\textit{UK}} & 38.4                            & 43.6                           & \multicolumn{1}{r|}{10.8}      & \textbf{41.3}                   & 46.9      & \multicolumn{1}{r|}{11.8}      & \multicolumn{1}{c}{--}          & \multicolumn{1}{c}{--}         & \multicolumn{1}{c|}{--}        & 28.9                            & 22.4                           & 12.9                           \\ \midrule
    \multicolumn{1}{r|}{\textit{HE}} & \textbf{53.7}                   & 7.9                            & \multicolumn{1}{r|}{37.8}      & 48.1                            & 14.9      & \multicolumn{1}{r|}{32.9}      & 50.5                            & 10.3                           & \multicolumn{1}{r|}{47.3}      & 46.6                            & 20.5                           & 24.5                           \\
    \multicolumn{1}{l|}{\textit{AR}} & 48.5                            & 43.7                           & \multicolumn{1}{r|}{16.1}      & 47.3                            & 48.3      & \multicolumn{1}{r|}{13.4}      & \textbf{49.8}                   & 38.5                           & \multicolumn{1}{r|}{19.0}      & 47.0                            & 49.4                           & 5.3                            \\ \midrule
    \multicolumn{1}{l|}{\textit{DE}} & 59.4                            & 12.5                           & \multicolumn{1}{r|}{12.5}      & {\ul \textbf{74.1}}             & 0.0       & \multicolumn{1}{r|}{30.2}      & {\ul 62.4}                            & 12.0                           & \multicolumn{1}{r|}{16.7}      & {\ul 48.6}                      & 34.5                           & 10.3                           \\ \bottomrule
    \end{tabular}%

  \caption{
    Performance of commercial MT systems on the \ds\ corpus on  all tested languages, categorized by their family: Spanish,
    French, Italian, Russian, Ukrainian, Hebrew, Arabic, and German.
    \emph{Acc} indicates overall gender accuracy (\% of instances the translation had the correct gender),
    $\Delta_G$ denotes the difference in performance ($F_1$ score) between masculine and feminine scores, and $\Delta_S$ is the difference in performance ($F_1$ score) between
    pro-stereotypical and anti-stereotypical gender role assignments (higher numbers in the two latter metrics indicate stronger biases).
    Numbers in bold indicate best accuracy for the language across MT systems (row), and underlined numbers indicate best accuracy for the MT system across languages (column).
    $^{*}$\aws\ does not have a trained model for English to Ukrainian.
    \label{tab:results}}
\end{table*}

\begin{table}[tb!]
  \centering
  \begin{tabular}{@{}lrrr@{}}
    \toprule
    & \multicolumn{1}{l}{\small{Acc}} & \multicolumn{1}{l}{$\Delta_G$} & \multicolumn{1}{l}{$\Delta_S$} \\ \midrule
    \textit{\small{FR \cite{ott2018scaling}}}          & 49.4                            & 2.6                            & 16.1                           \\
    \textit{\small{DE \cite{edunov2018understanding}}} & 52.5                            & 7.3                            & 8.4                            \\ \bottomrule
    \end{tabular}%
  \caption{Performance of recent state-of-the-art academic translation models from English to French and German.
  Metrics are the same as those in Table \ref{tab:results}.}
  \label{tab:results-sota}
  \end{table}

In this section, we briefly describe the MT systems and the target languages we use,
our main results, and their human validation.


\subsection{Experimental Setup}

\paragraph{MT systems}
We test six widely used MT models, representing the state of the art in both commercial and academic research:
(1) \goog,\footnote{\url{https://translate.google.com}}
(2) \bing,\footnote{\url{https://www.bing.com/translator}}
(3) \amazon,\footnote{\url{https://aws.amazon.com/translate}}
(4) \systran,\footnote{\url{http://www.systransoft.com}}
(5) the model of \citet{ott2018scaling}, which recently achieved the
best performance on English-to-French translation on the WMT'14 test set, and
(6) the model of \citet{edunov2018understanding}, the WMT'18 winner on English-to-German translation.
We query the online API for the first four commercial MT systems,
while for the latter two academic models we use the pretrained models provided by the Fairseq toolkit.\footnote{\url{https://github.com/pytorch/fairseq}}


\paragraph{Target languages and morphological analysis}
We selected a set of eight languages with grammatical gender
which exhibit a wide range of other linguistic properties (e.g., in terms of alphabet, word order, or grammar),
while still allowing for highly accurate automatic morphological analysis.
These languages belong to four different families:
(1) \textbf{Romance languages}: \textit{Spanish, French, and Italian}, all of which
have gendered noun-determiner agreement and spaCy morphological analysis support \cite{spacy2}.
(2) \textbf{Slavic languages} (Cyrillic alphabet): \textit{Russian and Ukrainian}, for which
we use the morphological analyzer developed by \newcite{pymorph2}.
(3) \textbf{Semitic languages}: \textit{Hebrew and Arabic}, each with a unique alphabet. For Hebrew, we use the
analyzer developed by \newcite{Adler2006AnUM}, while gender inflection in Arabic can be easily identified via the \emph{ta marbuta} character, which uniquely indicates feminine inflection.
(4) \textbf{Germanic languages}: \textit{German},
for which we use the morphological analyzer developed by \citet{Altinok2018DEMorphyGL}.



\pgfplotstableread[row sep=\\,col sep=&]{
  lang & pro & anti \\
  ES     & 67  & 46  \\
  FR     & 80 & 54  \\
  IT    & 52 & 30 \\
  RU   & 44 & 33 \\
  UK   & 46  & 35 \\
  HE      & 76  & 38 \\
  AR & 60 & 44 \\
  DE   & 69  & 57 \\
}\mydata

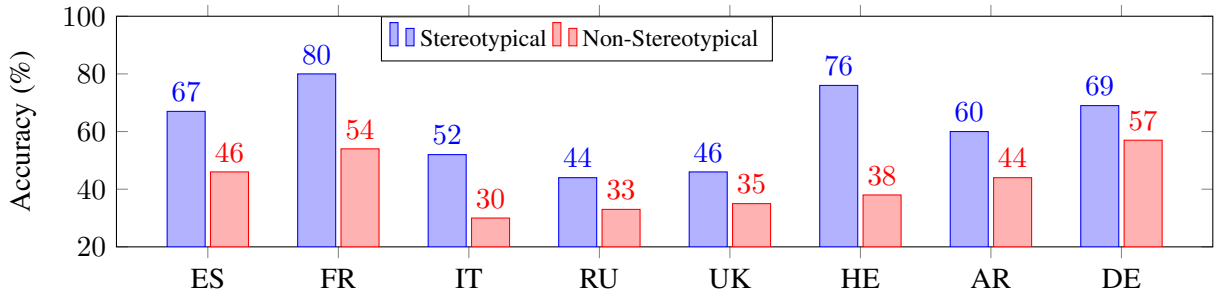
\begin{figure*}[t!]
\begin{tikzpicture}
  \begin{axis}[
      ybar,
      bar width=.5cm,
      width=\textwidth,
      height=.29\textwidth,
      legend style={at={(0.42,1)},
        anchor=north,legend columns=-1},
      symbolic x coords={ES,FR,IT,RU,UK,HE,AR,DE},
      xtick=data,
      nodes near coords,
      nodes near coords align={vertical},
      ymin=20, ymax=100,
      ylabel={Accuracy (\%)},
    ]
    \addplot table[x=lang,y=pro]{\mydata};
    \addplot table[x=lang,y=anti]{\mydata};
    \legend{{\small Stereotypical}, {\small Non-Stereotypical}}
  \end{axis}
\end{tikzpicture}
\caption{\goog's performance on gender translation on our tested languages.
  The performance on the stereotypical portion of \ds\ is consistently better than that on the non-stereotypical portion.
The other MT systems we tested display similar trends.}
\label{fig:google_acc}
\end{figure*}

\subsection{Results}
Our main findings are presented in Tables \ref{tab:results} and \ref{tab:results-sota}.
For each tested MT system and target language we compute three metrics
with respect to their ability to convey the correct gender in the target language.
Ultimately, our analyses indicate that all tested MT systems are indeed gender biased.

First, the overall system \emph{Accuracy}
is calculated by the percentage of instances in which the
translation preserved the gender of the entity from the original English sentence.
We find that most tested systems across eight tested languages perform
quite poorly on this metric.
The best performing model on each language often does not do much better than a random guess for
the correct inflection.
An exception to this rule is the translation
accuracies on German, where three out of four systems acheive their best performance.
This may be explained
by German's similarity to the English source language \cite{hawkins2015comparative}.

In Table~\ref{tab:results},  $\Delta_G$ denotes the difference in performance ($F_1$ score) between male and female translations.
Interestingly, all systems, except \bing\ on German, perform significantly better on male roles, which may stem from these being more frequent in the training set.

Perhaps most tellingly,
$\Delta_S$ measures the difference in performance ($F_1$ score) between stereotypical and non-stereotypical gender role assignments, as defined by \newcite{Zhao2018GenderBI} who use statistics provided by the US Department of Labor.\footnote{\url{https://www.bls.gov/cps/cpsaat11.htm}}
This metric shows that all tested systems have a significant and consistently better performance when presented with pro-stereotypical assignments (e.g., a female nurse), while
their performance deteriorates when translating anti-stereotypical roles (e.g., a male receptionist).
For instance, Figure \ref{fig:google_acc} depicts \goog\ absolute accuracies on stereotypical and non-stereotypical gender roles across all tested languages. Other tested systems show similar trends.

\begin{table}[tb!]
  \centering
  \begin{tabular}{@{}lrrr@{}}
    \toprule
    & \multicolumn{1}{c}{\small{Original}} & \multicolumn{1}{c}{\small{+Adj}} & \multicolumn{1}{c}{$\Delta$} \\ \midrule
    \textit{ES} & 53.1                                     & 63.5                                & +10.4                        \\
    \textit{RU} & 37.7                                     & 48.9                               & +11.2                       \\
    \textit{UK} & 38.4                                     & 42.9                                & +4.5                         \\ \bottomrule
  \end{tabular}
  \caption{Performance of \goog\ on Spanish, Russian, and Ukranian gender prediction accuracy (\% correct) on the original \ds\ corpus, versus a modified version of the dataset
    where we add sterotypical gender adjectives (see Section \ref{sec:adj}).}
  \label{tab:pretty}
  \end{table}

\subsection{Fighting Bias with Bias}
\label{sec:adj}
Finally, we tested whether we can affect the translations by
automatically creating a version of \ds\ with the adjectives ``handsome'' and ``pretty'' prepended to male and female entities, respectively.
For example, the sentence in Figure \ref{fig:flow} will be converted to: ``The \emph{pretty} doctor asked the nurse to help her in the operation''.
We are interested in evaluating whether this ``corrects'' the profession bias by mixing signals, e.g., while ``doctor'' biases towards a male translation,
``pretty'' tugs the translation towards a female inflection.
Our results show that this improved performance in some languages, significantly reducing bias in Spanish, Russian, and Ukrainian (see Table \ref{tab:pretty}).
Admittedly, this is impractical as a general debiasing scheme, since it assumes oracle coreference resolution,
yet it attests to the relation between coreference resolution and MT, and serves as a further indication of gender bias in MT.

\begin{table*}[]
  \centering
  \footnotesize
  \begin{tabular}{@{}p{0.31\textwidth}p{0.32\textwidth}p{0.31\textwidth}@{}}
    \toprule
    \textbf{Source}                                                               & \multicolumn{1}{l}{\textbf{[Target lang.] Predicted translation}}                             & \textbf{Phenomenon}                                   \\ \midrule

         The janitor does not like \neut{the baker} because \fem{she} always messes up the kitchen. & [ES] Al conserje no le gusta \masc{el panadero} porque \fem{ella} siempre desordena la cocina. & Biased translation, giving ``baker'' a male inflection, with a mismatched pronoun reference. \\ \midrule
         The janitor does not like \neut{the pretty baker} because \fem{she} always messes up the kitchen. & [ES] Al conserje no le gusta \fem{la panadera bonita} porque \fem{ella} siempre desordena la cocina. & Adding a stereotypically female adjective ``fixes'' the translation. \\ \midrule

    The counselor asked \neut{the guard} a few questions and praised \fem{her} for the good work. & [FR] Le conseiller a pos\'e quelques questions \`a \fem{la} \neut{garde} et \fem{l'a lou\'ee} pour le bon travail.  & French uses ``garde'' for both male and female guards, allowing for a more direct translation from English. \\
    \bottomrule
    \end{tabular}
  \caption{Examples of \goog's output for different sentences in the \ds\ corpus. Words in \masc{blue}, \fem{red}, and \neut{orange} indicate male, female and neutral entities,
    respectively.}
  \label{tab:examples}
\end{table*}

\subsection{Human Validation}
We estimate the accuracy of our gender bias evaluation method by randomly sampling 100 instances
of all translation systems and target languages, annotating each sample by two
target-language native speakers (resulting in 9,600 human annotations).
Each instance conformed to a format similar to
that used by our automatic gender detection algorithm:
human annotators were asked to mark the gender of an entity within a given target-language sentence.
(e.g., see ``el doctor'' as highlighted in the Spanish sentence in Figure \ref{fig:flow}).
By annotating at the sentence-level, we can account for both types of possible errors, i.e., alignment and
gender extraction.

We compare the sentence-level human annotations to the output of our
automatic method, and find that the levels of agreement
for all languages and systems were above 85\%, with an average agreement on 87\% of the annotations.
In comparison, human inter-annotator agreement was 90\%, due to noise introduced by several incoherent translations.

Our errors occur when language-specific idiosyncrasies introduce ambiguity to the morphological analysis.
For example, gender for certain words in Hebrew cannot be distinguished without diacritics (e.g., the male and female versions of the word ``baker'' are spelled identically),
and the contracted determiner in French and Italian (\emph{l'}) is used for both masculine and feminine nouns.
In addition, some languages have only male or female inflections for
professions which were stereotypically associated with one of the genders, for example ``sastre'' (tailor) in Spanish or  ``soldat'' (soldier) in French,
which do not have female inflections. See Table \ref{tab:examples} for detailed examples.

\section{Discussion}
\label{sec:related}
\paragraph{Related work}
This work is most related to several recent efforts which evaluate MT through the use of
\emph{challenge sets}.
Similarly to our use \ds, these works evaluate MT systems (either manually or automatically) on test sets
which are specially created to exhibit certain linguistic phenomena, thus going beyond the traditional BLEU metric \cite{papineni2002bleu}.
These include challenge sets for language-specific idiosyncrasies \cite{Isabelle2017ACS}, discourse phenomena \cite{Bawden2018EvaluatingDP},
pronoun translation \cite{DBLP:journals/corr/abs-1810-02268,Webster2018MindTG}, or coreference and multiword expressions \cite{burchardt2017linguistic}.

\paragraph{Limitations and future work}
While our work presents the first large-scale evaluation of gender bias in MT, it still suffers from certain limitations which could be addressed in follow up work.
First, like some of the challenge sets discussed above, \ds\ is composed of synthetic English source-side examples.
On the one hand, this allows for a controlled experiment environment, while, on the other hand,
this might introduce some artificial biases in our data and evaluation.
Ideally, \ds\ could be augmented with natural ``in the wild'' instances, with many source languages, all annotated with ground truth entity gender.
Second, similar to any medium size test set, it is clear that \ds\ serves only as a proxy estimation for the phenomenon of gender bias,
and would probably be easy to overfit. A larger annotated corpus can perhaps provide a better signal for training.
Finally, even though in Section \ref{sec:adj} we show a very rudimentary debiasing scheme which relies on oracle coreference system, it is clear
that this is not applicable in a real-world scenario.
While recent research has shown that getting rid of such biases may prove to be
very challenging \cite{Elazar:2018,Gonen2019LipstickOA}, we hope that this work will serve as a first step for developing more gender-balanced MT models.

\section{Conclusions}
We presented the first large-scale multilingual quantitative evidence for gender bias in MT, showing that on eight
diverse target languages, all four tested popular commercial systems and two recent state-of-the-art academic MT models are significantly prone
to translate based on gender stereotypes rather than more meaningful context.
Our data and code are publicly available at \link.

\section*{Acknowledgments}
We would like to thank Mark Yatskar, Iz Beltagy, Tim Dettmers, Ronan Le Bras, Kyle Richardson, Ariel and Claudia Stanovsky, and Paola Virga for many insightful discussions about the role gender plays in the languages evaluated in this work,
as well as the reviewers for their helpful comments.



\bibliography{acl2019}
\bibliographystyle{acl_natbib}

\end{document}